\documentclass[runningheads]{llncs}
\usepackage[T1]{fontenc}
\usepackage{graphicx}
\usepackage{amsmath, amssymb}
\usepackage[misc,geometry]{ifsym}
\begin{document}
\title{SynBT: High-quality Tumor Synthesis for Breast Tumor Segmentation by 3D Diffusion Model}

\author{Hongxu Yang\inst{1}(\Letter) \and 
Edina Timko\inst{2} \and 
Levente Lippenszky\inst{2} \and \\
Vanda Czipczer\inst{2} \and
Lehel Ferenczi\inst{2} 
}
\authorrunning{H. Yang et al.}
\institute{
Science \& Technology Org. AI \& ML, GE HealthCare, Eindhoven, Netherlands \and
Science \& Technology Org. AI \& ML, GE HealthCare, Budapest, Hungary\\
\email{Hongxu.Yang@gehealthcare.com}
}

\titlerunning{SynBT: Synthetic Breast Tumor}
%
%
%
\maketitle              
\begin{abstract}
Synthetic tumors in medical images offer controllable characteristics that facilitate the training of machine learning models, leading to an improved segmentation performance. However, the existing methods of tumor synthesis yield suboptimal performances when tumor occupies a large spatial volume, such as breast tumor segmentation in MRI with a large field-of-view (FOV), while commonly used tumor generation methods are based on small patches. In this paper, we propose a 3D medical diffusion model, called SynBT, to generate high-quality breast tumor (BT) in contrast-enhanced MRI images. The proposed model consists of a patch-to-volume autoencoder, which is able to compress the high-resolution MRIs into compact latent space, while preserving the resolution of volumes with large FOV. Using the obtained latent space feature vector, a mask-conditioned diffusion model is used to synthesize breast tumors within selected regions of breast tissue, resulting in realistic tumor appearances. We evaluated the proposed method for a tumor segmentation task, which demonstrated the proposed high-quality tumor synthesis method can facilitate the common segmentation models with performance improvement of 2-3\% Dice Score on a large public dataset, and therefore provides benefits for tumor segmentation in MRI images.

\end{abstract}

\section{Introduction}
Tumor synthesis provides the possibility of artificially generating the tumor examples in medical images~\cite{ChenR}. This method is valuable when there is insufficient number of training data for deep learning model, either without considerable amount of accurate voxel-level annotation, or with low variations in tumor characteristics. To train a robust deep learning model, training samples should typically cover a large characterization space, such as imaging protocol, patient demographics, tumor characteristics and image artifacts. These challenges drastically increase the efforts for data collection, cleaning and annotation, which increase the cost of model development targeting a single tumor type~\cite{ChenQ}.

In recent years, with the rapid achievements of image generation methods in deep learning, especially Generative Adversarial Networks (GAN) and Denoising Diffusion Probabilistic Model (DDPM)~\cite{GAN,DDPM}, medical images can be synthetically generated with high fidelity and valid clinical context~\cite{GAN,MAISI,MedDiff}. Nevertheless, these methods require a large amount of training images to generate volume-level medical images, making both data collection and model training expensive. In contrast, Chen et. al.~\cite{ChenQ} proposed a different strategy for image synthesis. The method~\cite{ChenQ} generates the tumor on a healthy image as the background, which reduces the effort of data collection and model training. However, this tumor synthesis method \cite{ChenQ} is designed only for early stage tumors in small patches, and is not feasible for breast tumor in MRI images with a typical volume size of $300^3$ voxels, therefore its generalization for different tumor types is compromised. 

In this work, we introduce a novel framework, called SynthBT to generate high-quality breast tumor in contrast-enhanced MRI images, which is able to boost the tumor segmentation performance on a public MRI dataset~\cite{MAMAMIA}. Specifically, the proposed generation framework for breast tumor segmentation consists of three major components: 1) A two-step patch-to-volume Vector Quantized Autoencoder (VQ-VAE)~\cite{VQVAE} training. Inspired by 3D MedDiffusion~\cite{MedDiffusion}, we implement a two-stage VQ-VAE training strategy to preserve the high-quality volumetric information of common MRI images. 2)  Training a diffusion model in the latent space, that is guided by the provided tumor masks.  The task of the generative model is to inpaint the removed tumor/healthy regions, with synthetic textures. 3) Training 3D segmentation models with large FOV, which is able to keep the semantic information for whole breast area, and to provide stable segmentation of the tumors. With controlling the tumor position and size, the diffusion model can produce a realistic tumor synthesis results, which leads to better segmentation performance on the considered open-sourced dataset.

The key contributions of this paper are two-fold. First, we introduce a high-quality tumor synthesis method for high-resolution breast MRI images. Extensive validation on the public dataset demonstrates that our method is effective and capable of enhancing performance in the targeted tumor segmentation tasks. Second, our work is the first to implement and validate a synthetic tumor generation framework on a large-scale public dataset with high variability in image quality for breast tumor segmentation in contrast MRI imaging, which shows the possible future directions for using synthetic data in AI model training for breast cancer applications.

\section{Methods}
\begin{figure}[htbp]
\centering{\includegraphics[width=11cm]{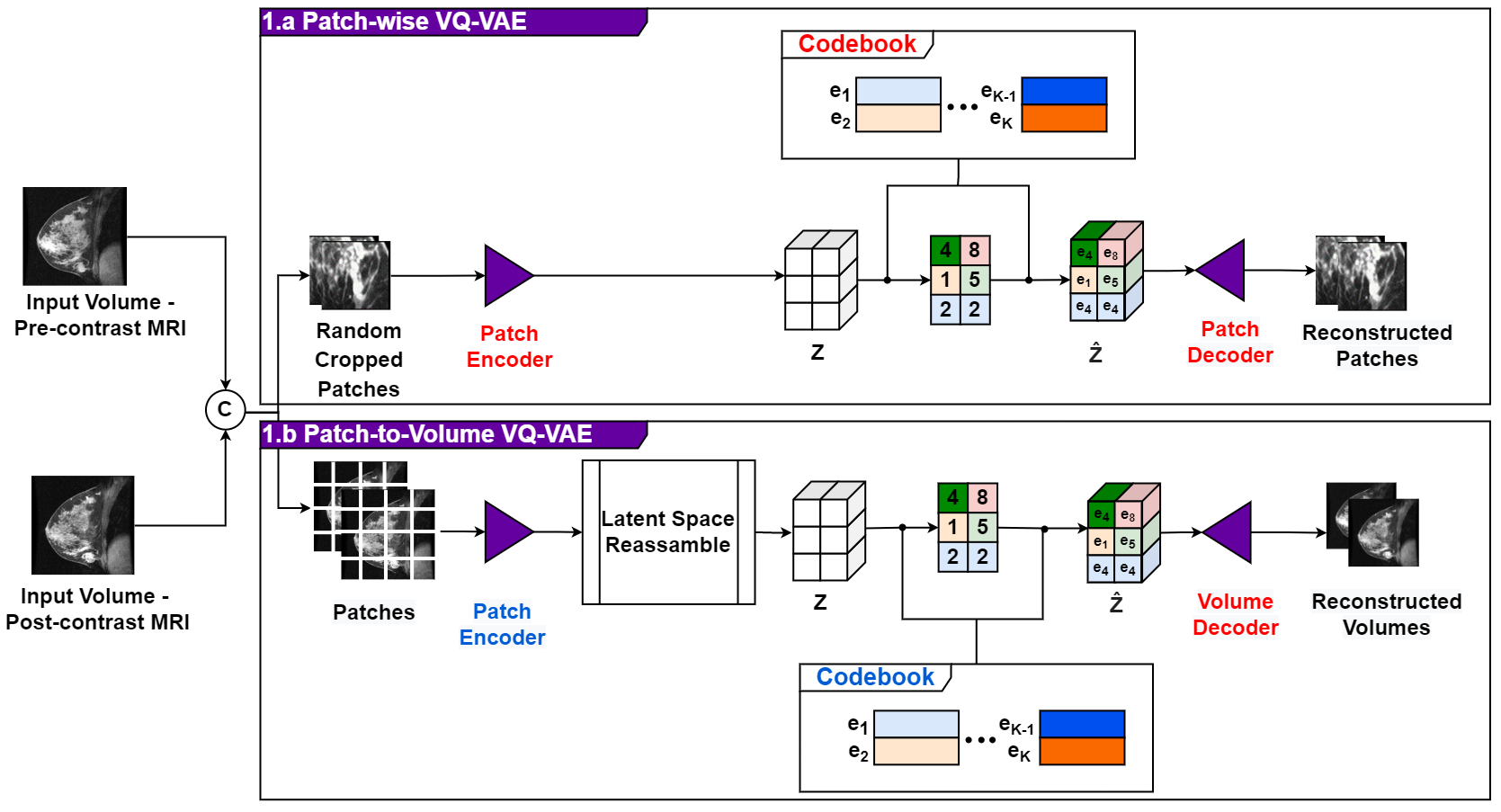}}
\caption{Patch-to-Volume VQ-VAE training has two steps. First, the VQ-VAE is trained on random sub-patch from the input MRI images (two sequences with one pre-contrast MRI and a random post-contrast MR image). The VQ-VAE is trained to compress and reconstruct the input information in the latent space with a quantization codebook. Second, the input volumes are divided into small patches, which are processed by patch-level encoder. The compressed latent vectors are reassembled to form the latent volume based on positional order, which are then processed by the codebook, a learnable quantization indexing method, and decoder. In the first stage, the VQ-VAE is fully optimized, while the second stage will only finetune the decoder for full volume with fixed encoder and codebook. Red: trainable. Blue: frozen.}
\label{vq_vae}
\end{figure}

\begin{figure}[ht!]
\centering{\includegraphics[width=11cm]{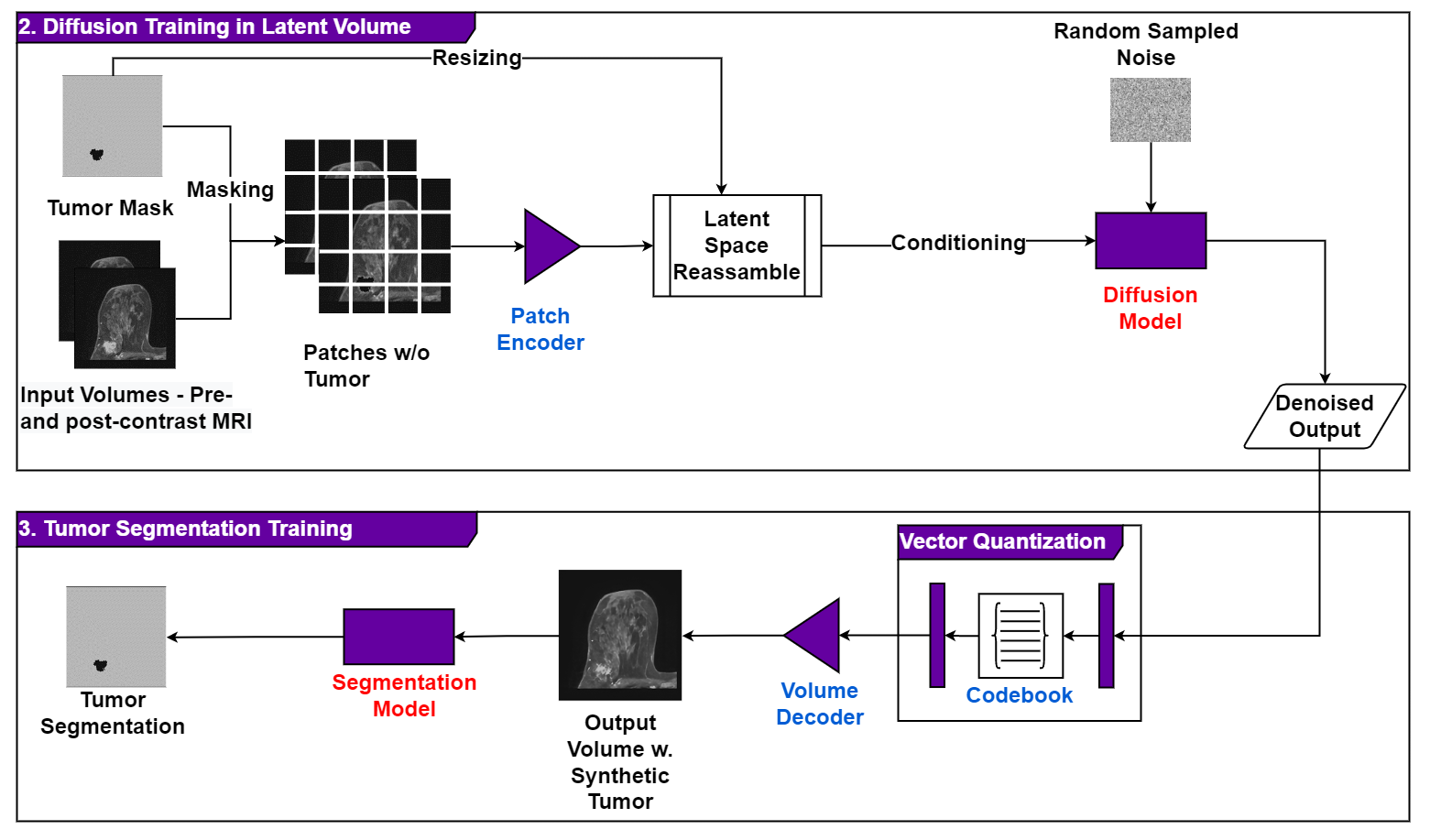}}
\caption{With trained Patch-to-Volume VQ-VAE, a latent space diffusion model is trained for synthetic tumor generation. In this stage, tumors in the volumes are masked by ground truth, which are processed by the encoder of VQ-VAE to generate latent space vectors without tumor. Tumors are synthetically generated by diffusion model from the conditioned binary mask. The final stage of the framework is the training of the segmentation models, which is optimized by real-tumor image together with synthetically generated tumor images. Red: trainable. Blue: frozen.}
\label{diff_seg}
\end{figure}

The proposed synthetic breast tumor generation and segmentation framework consists of three different models, as shown in Fig.~\ref{vq_vae} and Fig.~\ref{diff_seg}. The first stage is a two-steps VQ-VAE to efficiently compress the observed volumetric data into a latent space, while preserving the high-quality MRI volumes with large FOV. The second step is to train a latent space diffusion to synthetically generate the breast tumor based on the conditioning mask and the input volume. The final step is to train a semantic segmentation model to segment the breast tumor on MRI images by using synthetic images as supplementary data.
\subsection{Patch-to-Volume VQ-VAE}

The state-of-the-art tumor synthesis model~\cite{ChenQ} is designed to generate small tumor in patch-based segmentation model for CT images, which typically generates the target in a small patch of size $96^3$ voxels. Nevertheless, due to high-resolution of the breast DCE-MRI images, this small patch size fails to capture sufficient semantic information, which can compromise the segmentation performance. To address this, we consider using larger volumetric inputs. However, applying diffusion models directly in the image domain for large volumes incurs high computational costs and training instability. Therefore, diffusion in the latent space is preferred for 3D medical image generation ~\cite{MedDiff,MedDiffusion}. To obtain the latent space information, VQ-VAE is the most practical choice because of its highly efficient and low-cost design. Despite these advances, compressing a large volumetric data remains challenging for its memory consumption and training difficulties as the number of observed data points increases~\cite{MedDiffusion}. To overcome these limitations, 3D MedDiffusion~\cite{MedDiffusion} introduces a two-step region-to-global strategy for a high-quality VQ-VAE training by first learning local representations and then aggregating them into a global context. 

In our implementation of Patch-to-Volume VQ-VAE, we propose to use multi-sequence contrast-enhanced MRI as the input data. Specifically, the pre-contrast MRI $\mathbf{X}_{pre}$ and a post-contrast MRI $\mathbf{X}_{post}$ are combined as the input for the compression model: $\mathbf{X} = \text{Concat}(\mathbf{X}_{pre}, \mathbf{X}_{post}) \in \mathbb{R}^{H \times W \times D \times 2}$. This combined input is divided into small patches \( \mathbf{x}_i \in \mathbb{R}^{h \times w \times d \times 2} \), which are processed by the encoder \( E(\cdot) \) of the VQ-VAE to produce the latent vectors $\mathbf{z}_i = E(\mathbf{x}_i)$. These latent vectors are then reassembled based on their original grid position to form the latent volume $\mathbf{Z}$, which are then processed by element-wise quantization from a spatial-based codebook $\mathcal{C} = \{ \mathbf{e}_k \}_{k=1}^K$, where $\mathbf{K}$ is codebook size. The quantization operates each spatial code $z_i$ to its nearest vector $e_k$. The decoder $D(\cdot)$ reconstructs the volume from the quantized latent volume $\hat{ \mathbf{Z} }$, as $\hat{\mathbf{X}} = D(\hat{ \mathbf{Z} })$.

During the volume-level tuning, only the decoder is finetuned while keeping the encoder fixed to avoid the grid artifacts between different patches and to ensure a high-quality reconstruction. With this strategy, the training efforts for the encoder and decoder can be reduced, while encoder will only focus on the regional information extraction~\cite{MedDiffusion}.

\subsection{Latent Space Diffusion Model For Synthetic Tumor}
One of the commonly used approach for synthetic tumor generation in medical images is to synthetically embed the generated tumor regions onto healthy anatomical backgrounds, with carefully defined position, size, and texture. With limited clinical data under precise definition of tumor, we will not focus on the generation of background tissue, such as dense breast tissue. Instead, our method only models the foreground tumor with different varieties of its properties. 

In the proposed method, with a pair of tumor-present DCE-MRI volume and tumor mask $\mathbf{M}$, the diffusion model is trained to recover the tumor texture information by given the condition of $\mathbf{Z}_{\text{cond}} = \text{cat}(\mathbf{Z_{\text{masked}}}, \text{down}(\mathbf{M}))$, where $\text{down}(\cdot)$ is the down-sampling for mask to fit latent space size, while $\mathbf{Z_{\text{masked}}}$ is latent vector of the masked input volume (i.e., tumor removed). The forward diffusion process gradually adds Gaussian noise to the latent volume by $T$ steps:
\begin{equation}
q(\mathbf{z}_t \mid \mathbf{z}_{t-1}) = \mathcal{N}(\mathbf{z}_t; \sqrt{1 - \beta_t} \mathbf{z}_{t-1}, \beta_t \mathbf{I})
\end{equation}
where $\beta_t$ is a variance schedule that controls the amount of Gaussian noise added at each step. The reverse process is learned via a neural network $ \epsilon_\theta$, which predicts the noise added at each step, conditioned on the latent volume:
\begin{equation}
\mathbf{z}_{t-1} = \frac{1}{\sqrt{\alpha_t}} \left( \mathbf{z}_t - \frac{\beta_t}{\sqrt{1 - \bar{\alpha}_t}} \epsilon_\theta(\mathbf{z}_t, t, \mathbf{Z}_{\text{cond}}) \right) + \sqrt{\beta_t} \mathbf{n}, \quad \mathbf{n} \sim \mathcal{N}(0, \mathbf{I})
\end{equation}
where $\alpha_t = 1 - \beta_t$ and $\bar{\alpha}_t = \prod_{s=1}^t \alpha_s$ represent the cumulative product of noise retention factors over time \cite{DDPM}. This iterative denoising process reconstructs the tumor texture within the masked region in latent space guided by $\mathbf{Z}_{\text{cond}}$. Finally, the denoised latent volume $\hat{\mathbf{Z}}$ is passed through the VQ-VAE codebook and decoder to reconstruct the full-resolution image with the synthetic tumor.

\begin{figure}[ht!]
\centering{\includegraphics[width=11cm]{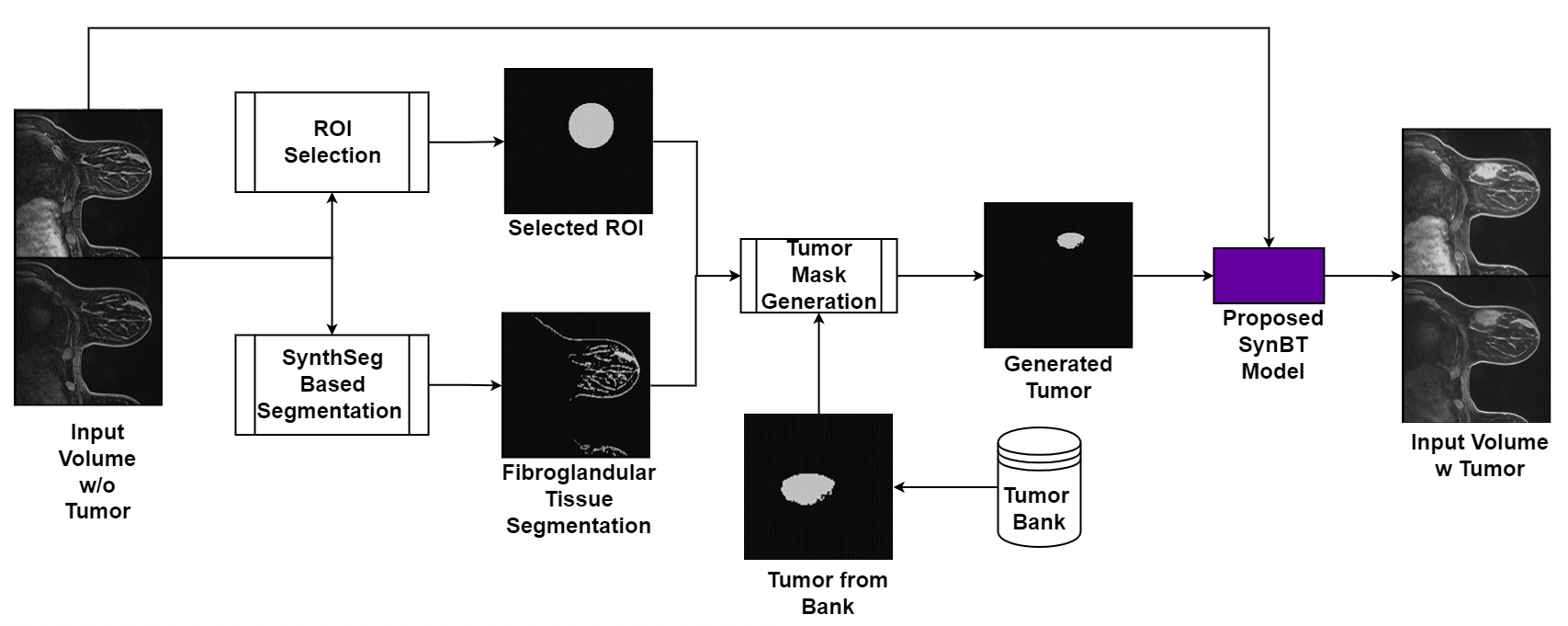}}
\caption{The synthetic tumor generation. With given input pre- and post-contrast MRIs, the tumor mask template is selected from the tumor template bank. The tumor mask template is processed to generate final tumor mask by combining with ROI selection and fibroglandular tissue segmentation, which are controlling the size, position, shape, etc. of the tumor. Finally, the synthetic tumor is generated in pre-/post-contrast MRIs. }
\label{syn_gen}
\end{figure}

\subsection{Segmentation Model Training with Synthetic Tumor}
The diffusion model is trained to generate the tumor texture by given tumor mask and background image, which is shown in Fig.~\ref{syn_gen}. In our proposal, the tumor mask is generated by a bank of template from real-tumor in contrast-enhanced MRI, which can be augmented by flipping, rotation, zoom, etc. to generate different possible tumor masks. In contrast, the tumor is positioned to closely align with the fibroglandular tissue within the breast. Its placement and size are determined based on both the tumor and breast dimensions, ensuring that the tumor is realistically located within the breast tissue. With the generated tumor mask from the rules defined above, and the properly selected background breast MRIs, the breast tumor can be synthesized in the pre- and post-MRI images with realistic appearances. Finally, the 3D segmentation model is jointly trained with real-tumor and on-the-fly generated synthetic tumor.
\section{Experimental Results}
\subsection{Materials and Implementation Details}
\subsubsection{Dataset:} We consider the large-scale, public MAMA-MIA dataset~\cite{MAMAMIA} for the experiment, which consists of DCE-MRI images with carefully annotated breast tumor. To conduct the experiment, 1200 images are used for training the VQ-VAE, diffusion model and segmentation models based on the official split. The remaining 306 images are further separated to 100 images for validation and 206 images for testing. The images are isotopically resampled to $1^3$ mm based on suggestion~\cite{MAMAMIA}. Specifically, due to multiple post-contrast sequences in the dataset, a random post-contrast MRI is selected to construct the dual-channel volume during the training, while the first post-contrast sequence is selected during the validation and testing phases. With a large amount of tumor masks in the training images, we collect these masks to construct the tumor bank for the synthetic tumor generation for segmentation tasks. In addition, the fibroglandular tissue masks from Duke dataset~\cite{duke} are used to train a tissue segmentation model based on SynthSeg pipeline~\cite{synthseg}, where the rest of the body parts are clustered by k-means clustering with the number of clusters randomly selected between 2 and 8. Specifically, 3D Duke fibroglandular masks are only used in training step of the SynthSeg model. Furthermore, the segmented tissue masks are post-processed to only select the most possible regions for placing a synthetic tumor. All the input volumes are randomly cropped from the resampled DCE-MRIs with size of $128\times192\times192\times2$ to have sufficient context and computational trade-off. 

\subsubsection{Implementation Details:} For VQ-VAE, patch size is $64^3\times2$ voxels, which is compressed by the VQ-VAE with compression rate of 4. The codebook has 8192 codes with dimensionality of 8. The initial convolutional channel is 32, and it's doubled as the patch resolution is reducing. The training mini-batch size is 8 with learning rate equals to $1\times10^{-4}$. The discriminator is started after 200,000 steps. To fine-tune the volume level reconstruction, the mini-batch size is set to 1 with learning rate of $1\times10^{-5}$ for 100,000 steps. In both trainings, the weight of the adversarial loss and the perceptual loss are set at 0.01 and 0.001~\cite{ChenQ,MedDiffusion}, respectively. To train a diffusion model, a specially designed UNet~\cite{ChenQ} is used for the generation of latent space. The denoising model is trained for 500,000 iterations with a learning rate of $1\times10^{-4}$. The total steps of diffusion scheduler is 1,000. During inference, the sampling is accelerated to 10 steps using the DDIM method~\cite{DDIM}. The segmentation on full volume is processed by sliding-window strategy with size of $128\times192\times192\times2$ voxels ($32\times48\times48\times8$ in latent space), where we consider three different segmentation models for segmentation based on MONAI implementation: U-Net, nnU-Net and SwinUNET~\cite{ChenQ,MONAI}. All segmentation models are trained to get the best validation score during the training for 300 epochs with real images, while the synthetic tumors are generated on-the-fly as supplementary dataset. In addition, data augmentations of flipping, rotation, intensity scaling and shifting are considered. The learning rate is $1\times10^{-4}$ for the training for mini-batch=1. All experiments were conducted on a DGX A100 40GB GPU. The segmentation performance is evaluated by Dice Score (DSC), 95\% Hausdorff Distance (95HD) and mean surface distance (MSD) using MedPy.
\subsection{Synthetic Tumor and Segmentation Performance}
\begin{figure}[ht!]
\centering{\includegraphics[width=11cm]{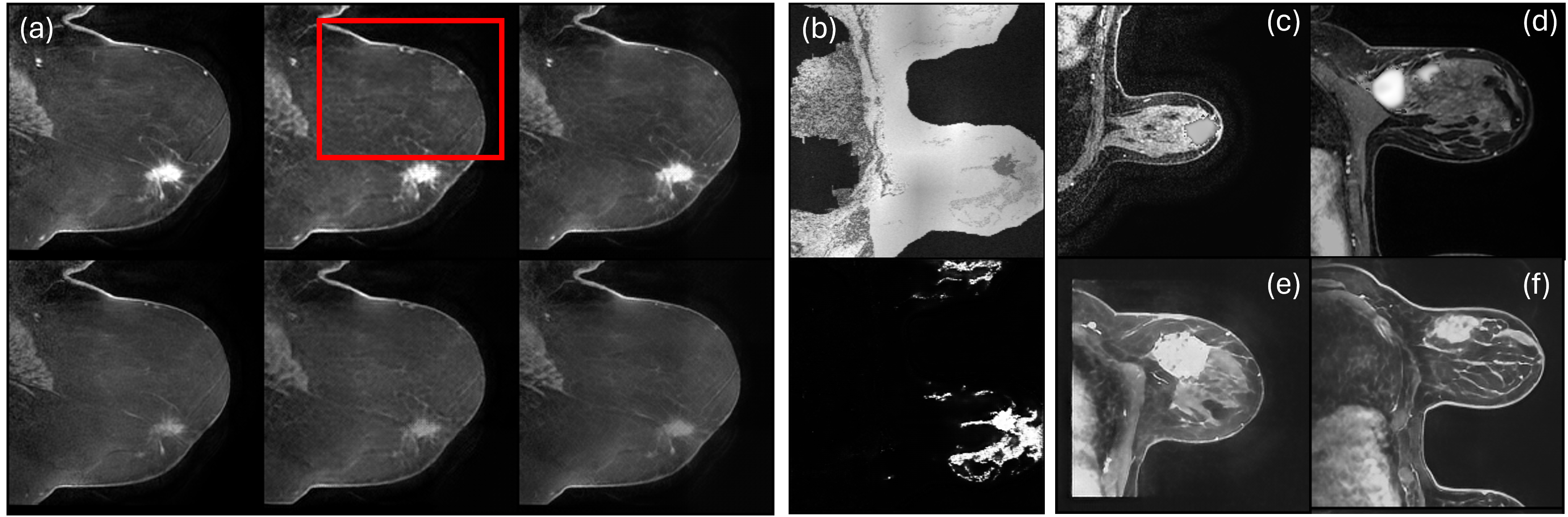}}
\caption{(a) Left to right: real image, patch-only VQ-VAE output and Patch-to-Volume VQ-VAE outputs. Top row is post-contrast MRI, while bottom row is pre-contrast MRI. Grid effect can be observed in mid-column images for patch-only VQ-VAE. (b) SynthSeg based method for fibroglandular segmentation. Top: fake image from masks by random Gaussian Mixture Model, bottom: fibroglandular tissue as mask. (c-d) Synthetic tumor from SynTumor. (e-f) Synthetic tumor from SynBT.}
\label{example}
\end{figure}

\begin{table}[ht!]
\centering
\caption{Performance comparisons of segmentation results by using Dice Score (DSC), 95\% Hausdorff Distance (95HD) and mean surface distance (MSD) on testing volumes.}
\begin{tabular}{l|c|c|c}
\hline
Method            & DSC (\%) $\uparrow$ & 95HD (mm) $\downarrow$ & MSD (mm) $\downarrow$ \\ \hline
patch-UNet         & 0.3907   & 136.83    & 75.80    \\ 
Volume-UNet        & 0.7220   & 60.05     & 23.90    \\ 
SynTumor-UNet        & 0.6980   &  63.28    &  20.34   \\ 
SynBT-UNet       & \bf{0.7326}   & \bf{53.79}     & \bf{17.71}    \\ \hline
patch-nnUNet       & 0.3758   & 153.15    & 81.81    \\ 
Volume-nnUNet      & 0.7463   & 30.77     & 12.48    \\ 
SynTumor-nnUNet        &  0.7266  &  37.73    &   16.24  \\ 
SynBT-nnUNet     & \bf{0.7629}   & \bf{29.26}     & \bf{10.72}    \\ \hline
patch-SwinUNETR    & 0.4667   & 118.41    & 64.37    \\ 
Volume-SwinUNETR   & 0.7278   & 45.25     & 16.65    \\ 
SynTumor-SwinUNETR        &  0.6750  &   65.11   &   29.22  \\ 
SynBT-SwinUNETR  & \bf{0.7580}   & \bf{33.99}     & \bf{11.60}    \\ \hline
\end{tabular}
\label{performance}
\end{table}

The performance and ablation study of SynBT on the segmentation task are summarized in Table~\ref{performance}. Specifically, three different commonly considered segmentation models are reported in the table. The patch-* represents the segmentation models that are trained on small patches of $64^3$ voxels, which indicates the importance of having a large field-of-view (FOV) for tumor segmentation in high-resolution DCE-MRI images. As one can observe, the patch-based segmentation model cannot handle the complex context of the breast region. In contrast, using a larger volume size can capture richer semantic information, leading to improved segmentation performance. This highlights the importance of generating high-quality synthetic data to support such methods. To compare the different methods of tumor texture generation, instead of considering the diffusion-based method, we also compare the state-of-the-art analytical tumor synthesis model, SynTumor~\cite{LabelFree}. In our experiments, we found that it has a worse performance for the breast tumor synthesis, as its texture generator is not capable of generating realistic breast tumors with contrast enhanced appearance, as shown in Fig.~\ref{example}. Finally, the proposed synthetic tumor generation method (SynBT-) consistently enhances the performance across various segmentation models and metrics, demonstrating its effectiveness in improving tumor segmentation.

\section{Conclusions and Discussions}
In this work, we propose a synthetic breast tumor generation workflow for DCE-MRI imaging, which demonstrates its potential to improve the performance of segmentation models. The method includes several key steps for segmenting breast cancer, namely Patch-to-Volume image compression, mask-guided tumor generation and tumor segmentation in DCE-MRI. The method is validated on a large-scale public dataset. The validation highlights its ability to generate realistic synthetic tumors and shows improvement for segmentation performance. 

Nevertheless, there are still several limitations that could be addressed in future work. For instance, one direction could be the work for more precise placement of the tumor, due to its challenging nature in breast cancer. Our work is based on the assumption that the tumors generally exist in randomized location at the fibrograndular tissue of the breast. In clinical practice, this position information can be more precisely defined, which could lead to a potential increase in the semantic definition of the synthetic tumor. Additionally, it would be beneficial to improve the tissue segmentation model, as it requires high efforts from post-processing side. Moreover, the diffusion based model has limitations of correctly controlling the texture with desired behavior, such as controlling the tumor categories, dedicated research could be conducted to improve the controllability of the tumor texture. Finally, dedicated quantitative synthetic image quality analysis should be conducted in the future, such as using FID, SSIM and similarity of tumor Radiomics features.

\begin{credits}
\subsubsection{\ackname} This research was supported by SYNTHIA, a project supported by the Innovative Health Initiative Joint Undertaking (IHI JU) under grant agreement No 101172872. The JU receives support from the European Union's Horizon Europe research and innovation programme. Disclaimer: Funded by the European Union, the private members, and those contributing partners of the IHI JU.  The views and conclusions expressed in this paper are those of the author(s) only and do not necessarily reflect those of the aforementioned parties. Neither of the aforementioned parties can be held responsible for them.

\subsubsection{\discintname}
This work was conducted by the authors as part of their full-time employment at GE HealthCare.
\end{credits}

\bibliographystyle{splncs04}
\bibliography{Paper-DeepBreath2025-0012}

\end{document}